\documentclass[10pt, conference, compsocconf]{IEEEtran}

\usepackage{xcolor}
\usepackage{longtable}
\usepackage{bm}
\usepackage{booktabs}
\usepackage{footnote}
\usepackage{zhnumber} 		
\usepackage{authblk}
\usepackage{amsfonts}
\usepackage{graphics}
\usepackage{verbatim}
\usepackage{lipsum}
\usepackage{subfigure}
\usepackage{verbatim}
\usepackage{mathrsfs}
\usepackage{enumitem}
\usepackage{tabularx}
\usepackage{amsmath}
\usepackage{stfloats}

\usepackage{algorithm} 
\usepackage{algorithmicx}  
\usepackage{algpseudocode}   
\floatname{algorithm}{Algorithm}

\usepackage{latexsym}
\usepackage{url}
\usepackage{graphicx}

\newtheorem{definition}{Definition}

\begin{document}

\bibliographystyle{IEEEtran}

%
\title{User Intention Recognition and Requirement Elicitation Method for Conversational AI Services}

\author{Junrui Tian\IEEEauthorrefmark{1},
		Zhiying Tu\IEEEauthorrefmark{2}, Zhongjie Wang\IEEEauthorrefmark{3},
		Xiaofei Xu\IEEEauthorrefmark{4} and Min Liu\IEEEauthorrefmark{5} \\
Harbin Institute of Technology, China\\
	Email: \IEEEauthorrefmark{1}junruit97@gmail.com,
	\IEEEauthorrefmark{2}tzy\_hit@hit.edu.cn,
	\IEEEauthorrefmark{3}rainy@hit.edu.cn,
	\IEEEauthorrefmark{4}xiaofei@hit.edu.cn,
	\IEEEauthorrefmark{5}1549211994@qq.com}

\maketitle

\begin{abstract}
In recent years, chat-bot has become a new type of intelligent terminal to guide users to consume services. However, it is criticized most that the services it provides are not what users expect or most expect. This defect mostly dues to two problems, one is that the incompleteness and uncertainty of user's requirement expression caused by the information asymmetry, the other is that the diversity of service resources leads to the difficulty of service selection. Conversational bot is a typical mesh device, so the guided multi-rounds Q$\&$A is the most effective way to elicit user requirements. Obviously, complex Q$\&$A with too many rounds is boring and always leads to bad user experience. Therefore, we aim to obtain user requirements as accurately as possible in as few rounds as possible. To achieve this, a user intention recognition method based on Knowledge Graph (KG) was developed for fuzzy requirement inference, and a requirement elicitation method based on Granular Computing was proposed for dialog policy generation. Experimental results show that these two methods can effectively reduce the number of conversation rounds, and can quickly and accurately identify the user intention.
\end{abstract}


%
\IEEEpeerreviewmaketitle
\vspace{-1.5ex}
\section{Introduction}\label{sec:intro}
In recent years, Apple's Siri, Microsoft Cortana, and other service products have become more and more popular. Conversational AI bot, such as Intelligent Voice AI Assistant, has been trained to understand voice commands and complete tasks for users in various application scenarios. The convenience of conversational AI bot makes the cognitive service computing system an inevitable trend in the future. We had developed this kind of Cognitive Service Robot, to help users to find appropriate services and construct coarse-grained service solutions \cite{xu2019sbot}. According to the method referred in above, we perform domain validation predictions for user requirements. During the conversational AI bot development, we found that massive services with various functional and non-functional attributes make it very difficult for cognitive service to select the user expected services automatically and accurately.This problem requires the user requirements, including dominant or implicit, can be recognized automatically.

Understanding and feedback are two basic abilities for bots. However, they are great challenges for bots to achieve intelligence. Currently, there are three kinds of bots, Q$\&$A bot, Task bot, and chat-bot. Most of the Q$\&$A bot prefers to deal with single round conversations, rather than the complex ones with context. Task bot are specialized for one domain related mission, such as navigation, or any kinds of consulting. They might reject to provide service once the request beyond their ability, which is given by domain prior knowledge. Chat-bots are more like a pet, only responsible for entertainment-related matters.

\begin{figure}[htbp]
\vspace{-3ex}
\centering
\includegraphics[height=4.0cm,width=8.0cm]{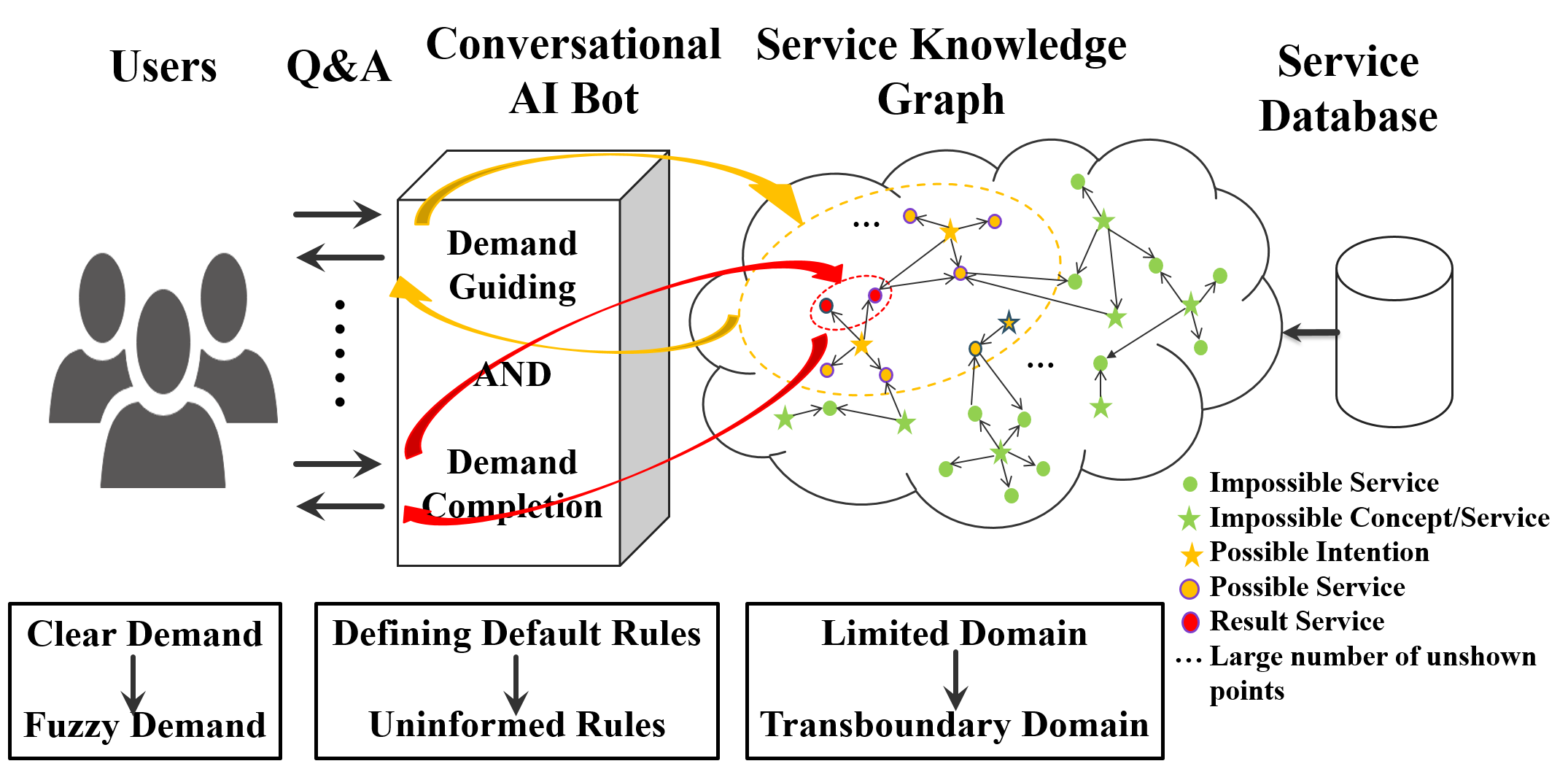}
\caption{Challenge and Contribution}
\label{fig:intro}
\vspace{-3ex}
\end{figure}

Conversational AI bot can be considered as a cognitive mediator to lead users to the service world. The two methods proposed in this paper can help bots to adapt to more complex scenarios and deal with the intention contained in the user's fuzzy and uncertain requirement expressions. Their accurate recognition results would further support bot to make service recommendation decisions by selecting different services from various domains (as shown in Fig. \ref{fig:intro}). In addition to the above tasks, these two methods also need to contribute on the following breakthroughs. So, the human-computer interaction of bot would be more natural and intelligent. Firstly, to minimize the formatting of system input, especially rule-based input. Secondly, to make the method of user intention understanding more flexible and intelligent. Last but not least, not to be single specific domain. 
Different from the traditional dialogue mode, when the user needs a housekeeping cleaning service, he doesn't need to give the command ``please make an appointment for housekeeping". Instead, he may inadvertently say, ``the kitchen is a little dirty". The bot should automatically infer the user's intention, and then clear the details of the service, such as \textit{price} and \textit{time}, through several rounds of inquiry.


Generally, the multi-round conversation for eliciting user requirements is led by a knowledge graph based inference rather than the predefined domain rules. This knowledge graph is constructed based on the description of services from several domains by using our previous work \cite{tu2019crowdsourcing}. The domain projection method mentioned in this work and the \textit{Trans} series methods \cite{wang2017knowledge} are used to infer the potential user intention, further on to define the searching space. And then, a method based on granular computing is proposed to auto-decide the dialogue strategy (the next-round selection) within this search space. Finally, the candidate service solutions for recommendation would sort according to their QoS evaluation. 

This paper would be organized as follows: Section \uppercase\expandafter{\romannumeral2} introduces the state of the arts in related fields. Section \uppercase\expandafter{\romannumeral3} discusses the motivation and problem definition. The overall problem, including the basic idea, the overall process and the overall structure, will be analyzed and defined there. Section \uppercase\expandafter{\romannumeral4} introduces our method and data model to solve the problem from different aspects. Section \uppercase\expandafter{\romannumeral5} demonstrates the process and results of the experiments and the performance comparison with existing awareness service systems. Finally, the paper makes the conclusions and open future work for discussion in Section VI.

\vspace{-1ex}
\section{Related Work}
\vspace{-1ex}
\subsection{The status of chat-bot.}
\vspace{-1ex}
Early chat-bots generally adopted a rule-based Q$\&$A bot bases on a semantic template called an expert system. The system finds the best answer through the template matching paradigm. In recent years, the structure of the Q$\&$A system has also transformed from a traditional template-based method to a knowledge-based approach. The system can use search, logical reasoning, and other ways to find answers to user questions. But it is limited to the scale of the knowledge graph when input requests are outside the scope of the knowledge graph, the stability of the answer drops distinctly. 

\begin{table*}[hb]
\scriptsize
\vspace{-5.0ex}
\centering
\caption{The Comparison of The Current Status of the Bot}
\label{tab:compare_chatbot}
\begin{tabular}{ccccc}
\toprule
\textbf{Type of Bot}&\textbf{Attention}&\textbf{State}&\textbf{Benifit}&\textbf{Weakness}\\ 
\midrule
\begin{tabular}[c]{@{}c@{}}QA Bot based on \\ Reading Comprehension\end{tabular} & \begin{tabular}[c]{@{}c@{}}Confirm User Question\\ and Answer\end{tabular} &  \begin{tabular}[c]{@{}c@{}}Understand \\ User Intention \end{tabular} & \begin{tabular}[c]{@{}c@{}}Pinpoint the answer\\ to the question;\\ Answer highly relevant;\end{tabular} & \begin{tabular}[c]{@{}c@{}}Single dialogue;\\ Context free;\end{tabular} \\ \hline
\begin{tabular}[c]{@{}c@{}}Task Bot\\ based Dialogue System\end{tabular} & \begin{tabular}[c]{@{}c@{}}Take action\\ and Extract keywords\\ Based Dialogue\end{tabular} & Clear the propose & \begin{tabular}[c]{@{}c@{}}Pinpoint to the domain;\\ High probability of success\\ in each round;\\ Multi-dialogue;\end{tabular} & \begin{tabular}[c]{@{}c@{}}Resolving for\\ a specific\\ scenario\end{tabular} \\ \hline
Chat-bots & Answer and Response & \begin{tabular}[c]{@{}c@{}}The History of\\ Communication\end{tabular} & \begin{tabular}[c]{@{}c@{}}Natural Interaction\\ without limited domain;\\ Multi-dialogue;\end{tabular} & \begin{tabular}[c]{@{}c@{}}Unable to\\ solve problem\end{tabular} \\ \hline
\begin{tabular}[c]{@{}c@{}}KG Conversational AI\\ bot based on\\ Recommendation \\for Cognitive Services\\ Method (Our Bot)\end{tabular} & \begin{tabular}[c]{@{}c@{}}Understand User \\ requirement and Match\\ Service for Feedback\end{tabular} & \begin{tabular}[c]{@{}c@{}}Clear the requirement\\ and filter the possible\\ service base on\\ Multi-dialog System\end{tabular} & \begin{tabular}[c]{@{}c@{}}Accept unlimited input;\\ Pinpoint the dialog domain;\\ Clear the condition by Multi-dialogue;\\ Reasoning based on KG;\\ Return the proper solutions;\end{tabular} & \begin{tabular}[c]{@{}c@{}}Lack Of\\ the Personal\\ portrait\end{tabular} \\ 
\hline
\end{tabular}
\vspace{-3ex}
\end{table*}

Chat-bots used information retrieval technologies to achieve the best match between questions and answers since 1995. They are not prone to grammatical errors, but they may not be able to deal with scenes without pre-defined. After receiving the user data, bots use a specific method to create a sentence automatically \cite{ritter2011data}\cite{shang2015neural}. The benefit is that the bot can cover user questions on any topic as the response. But the disadvantage is that the quality of the response sentence generated may have problems. For example, the statement may not be fluent or has syntax errors and other low-level errors. The chat-bots can serve in the open domain or limited domain. In the open-domain environment, the user can chat with the bot about any content belonging to the open-domain category  (still countable domain). In the limited field, the user can barely talk about the dialogues that are not preset within this specific domain.

As an open-source bot, ALICE built according to rules that receive input and generate output \cite{cho2007emotional}, but there are no automatic statements and questions and no response to user input. Google Now is a smart personal assistant software that answers questions and provides suggestions through a series of web services. It also predicts what information may be needed based on the user's past search habits. This paper gives the comparison of various types of robots shows in TABLE  \ref{tab:compare_chatbot}.
\vspace{-1ex}
\subsection{Requirement acquisition and analysis}
\vspace{-1ex}
Requirement acquisition and analysis is an essential part of accomplishing project tasks. Traditional methods use user records to identify intentions. Letizia and Lieberman \cite{Lieberman1995An} proposed to represent documents of interest to users as keyword vectors, and calculated each keyword weight to establish a user demand model. Li et al \cite{li2014weakly} used weakly supervised learning methods to extract user-related information from Twitter social data. Venkatesan \cite{Venkatesan2015Mining} records user requirements by analyzing the user's behavior log. Srinivasan and Batri \cite{Srinivasan2013Reducing} conduct user requirements analysis by users' search records on the server. 

The traditional chat-bots obtain the query intention by directly matching the word list \cite{higashinaka2006incorporating}. At the same time, it can accurately solve the high-frequency words by adding categories that are relatively simple and have relatively concentrated query patterns. However, it requires more human participation, which is challenging to automate. 

Ontology analysis is also a common method in traditional requirements analysis methods. In 2006, Kaiya and Saeki \cite{Kaiya2006Using} used ontology to build a domain knowledge base for requirement analysis. 
(For example, \textit{``today's air ticket price from $a$ to $b$"} can be converted into $[location$] to $[location][date][bus/air/train ticket]$ price.)
This method of intention recognition by rules has better recognition accuracy for the requirement with strong regularity and can extract accurate information. However, the process of discovering and formulating rules also requires more human participation.
Zhang and Wallace regard intention recognition as a classification problem and define different categories of requirement intention according to the characteristics of vertical products, and common words for each intention category can be counted \cite{zhang2015sensitivity}. For the requirement input by the user, the probability of each intention is calculated according to the statistical classification model, and finally, the intention of the requirement is given.
Google proposed the Bidirectional Encoder Representation from Transformers Model (BERT), which significantly improved the ability to identify user intention in 2018 \cite{devlin2018bert}. The BERT model is used to solve the Q$\&$A and service recommendation problems based on the KG.

\vspace{-1ex}
\section{Motivation and Problem Definition}
\vspace{-0.5ex}
\subsection{Motivation}\label{subsec:motivition}
\vspace{-1ex}
In general, this study aims at proposing a human-like man-machine dialogue method. This method can lead users to express their requirements completely and accurately, step by step. Currently, most of the existing chat-bots work quite well with the requirement expression in a fixed format or trigger. However, this is not a human style, in which every single expression in a conversation could be full of metaphor, implication, and personalization. This would cause the incompleteness, diversity, or implication of the user requirement proposition. It means uncertainty that brings a huge challenge to the intention understanding of the bots. The partial semantic description and various expressions, especially for specific nouns and local dialects, extremely affect the accuracy of intention identification. Thus, it makes the original expression of user requirements difficult to be translated into machine understandable service objectives.

Artificial design of semantic slots is the most commonly used way for intention understanding of chat-bots. Traditional Q\&A uses feature extraction and matching method to realize requirements understanding and identify requirements domain, which is based on rules, word embedding, or classifier (such as Bayes). However, it has a strong dependence on domain knowledge, which makes it difficult to switch between different domains in conversation semantic space and action space. Besides, most of the existing question understanding methods are focus on the questions of single sentence form, which also relies on a specific sentence structure. Therefore, to not only identify \textit{person}, \textit{location}, \textit{organization}, and \textit{date} but identify also much more fine-grained entity types, this paper use the \textbf{BERT} model to enhance the ability of recognition. 
The traditional Q$\&$A modes are always suffered from problems like inaccurate information retrieval, redundant Q$\&$A information error, etc. Although big data and deep learning methods have greatly improved their accuracy, the number of samples substantially limits the personalization of the answers. However, different people will not always have the same requirement. Therefore, guided multi-round Q$\&$A should be the most suitable way for requirements elicitation. This requires the method of requirement pruning to generate dialogue strategy.

The decision tree algorithm is one of the commonly used pruning methods. Typically, when the number of candidate service is small, this algorithm can correctly and efficiently classify the services. However, once the number of candidate service become larger and larger, the tree would become more complex with tremendous nodes. Then, this algorithm would not be possible to perform well as expected, no matter in accuracy or efficiency. While the granular computing \cite{Lin2003Granular} can deal with large-scale problems, many reasoning algorithms combined with rough set theory and granular computing theory, and form multi-granular and multilevel analysis and processing methods. Granular computing theory can perform granular analysis on the domain information represented by big data, and determine the number of possible granularity levels. The results of the analysis and its quality can affect the accuracy and efficiency of dialogue strategy generation. This paper selects suitable multi-granularity modeling for specific data to achieve support for specific GrC models that can better perform data analysis.

\vspace{-1ex}
\subsection{Problem Definition}
\vspace{-1ex}
\begin{definition}[Service Knowledge Graph]\label{KG} Knowledge graph is a structured semantic knowledge base. Service knowledge graph $G\in(E,R,S)$, and $E= \{e_{1},e_{2},\cdots,e_{\vert E\vert}\}$ is the set of the entities in knowledge base, which includes $\vert E \vert$ different entities. Entities includes service entity nodes $sn_{i}$ and attribute nodes $Attribute_{j}$.
$R$ = $\{r_{1},r_{2},\cdots,r_{\vert R \vert} \}$ is the set of the relationships in knowledge base, which includes $\vert R\vert$ different relationships. Realtionships in service KG are the labels of attributes
$S \subseteq E \times R \times E$ represents the set of triples in the knowledge base. The semantic is ``The $r_{k}$ of $sn_{k}$ is $Attribute_{k}$" ($k$ is any value).
\end{definition}
\begin{definition}[User Initial Intention]\label{def:intrec} System accepts an initial user fuzzy requirement as input $S$. And the paper can get an  analysis result $D$ of user intention, where $D$ includes lots of requirement concerns $D_{1},D_{2},\cdots,D_{i},\cdots $.
For every $D_{i}$, there are some restrict sets $H_{ij}$. Restrict set is a set of a label with some attributes. The label is the identified entity type by BERT model. The attributes of the label are the constraint conditions that the user proposes. The structure of $H_{ij}$ and $D_{i}$ are shown as below:
\vspace{-1ex}
\begin{equation} \label{HIJ}
D_{i} =  \left\{ H_{i1},H_{i2},\cdots,H_{ij},\cdots \right\}, \forall i,j > 0 
\end{equation}
\begin{equation}
\vspace{-1.5ex}
\label{HASHSet}
H_{ij} =  label : \left\{ Attribute_{1},Attribute_{2},... \right\}
\end{equation}
\end{definition}
\begin{definition}[Requirement Pruning Strategy]\label{PRUNE} 
This strategy aims at finding the mutually exclusive service attributes for individual requirement inquiry. Thus, all services need to be optimally clustered into $C=\left\{C_{1}, C_{2}, C_{3}, \cdots, C_{k}\right\}$ according to these different attributes. The optimization objective function of clustering algorithm is defined as equation \ref{fa}. $\mu_{ij}$ is the membership degree of service $sn_{j}$ and cluster $i$, as defined in equation \ref{35}. $m$ is a weighted value; $d_{ij}$ is the distance between the attribute vector of the service $sn_{j}$ and the cluster $i$ vector, which is recorded as $\sqrt{\sum_{t}^{N_{a}}\left(y_{t}-m_{it}\right)^{2}}$. Where $y_{t}$ and $m_{it}$ are the value mapping of the service $sn_t$ and centroid point $m_{it}$ on attribute $t$ ($m_{i}$ is the clustering center vector), $N_{a}$ is the number of attributes.
\begin{equation}\label{35}
\mu_{i j}=\left(\sum_{t=1}^{c}\left(\frac{d_{i j}}{d_{t j}}\right)^{\frac{2}{m-1}}\right)^{-1}
\end{equation}
\begin{equation}\label{fa}
Goal=\sum_{j=1}^{n} \sum_{i=1}^{c}\left(\mu_{i j}\right)^{m}\left(d_{i j}\right)^{2}
\end{equation}

\end{definition}


\vspace{-1.5ex}
\section{Our Method of Solving Plan}
\vspace{-0.5ex}
\subsection{Overview of the Framework}
\begin{figure}[htbp]
\centering
\includegraphics[width=0.9\linewidth]{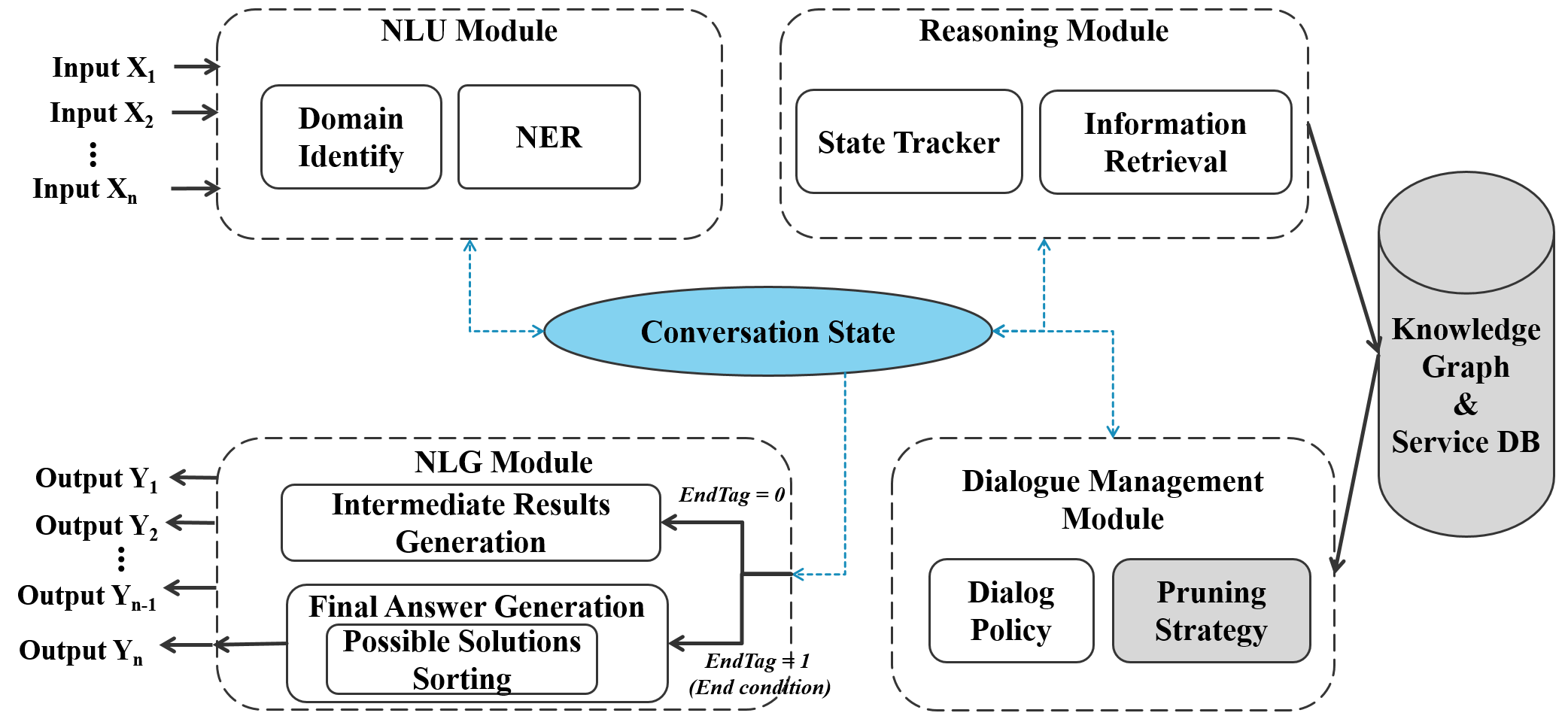}
\caption{Our framework of system}
\label{module2}
\vspace{-1ex}
\end{figure}

\vspace{-0.5ex}
As shown in Fig. \ref{module2}, the framework proposed in this paper consists of four major modules, NLU (Nature Language Understanding) module, Reasoning Module (RM), Dialogue Management Module (DMM), and NLG (Nature Language Generation) module. NLU identifies the domain that user requirement belongs to, and the intentions implied in the user expression. RM decides what the candidate services are. DMM leads to a conversation based on the dialog policy, which is generated by an offline module, pruning strategy. NLG prepares the reply for each round of the conversation.


The framework proposed in this paper contains context information and conversational state management. It can enable the system to lead the conversation and identify the user complete demand proposition. Meanwhile, this process should end before the user gets bored. This framework adopts the fuzzy method to achieve the auto-execution of this process. Instead of the predefined logic rules of requirement slot filling, a granular computing method is used to cluster the human services data to auto-decide the order of requirement pruning and generate the feedback of each round in the conversation. In every round of the conversation, this framework uses a corresponding ``conversational state” to represent the progress of the conversation. KG is used for implied intention reasoning and answer generation. Table \ref{process} demonstrates an example of conversation processing based on this framework.

\vspace{-2ex}
\begin{table}[htbp]
\scriptsize
\centering
\caption{Example of the first round system process based on KG}
\label{process}
\begin{tabular}{ccc}
\hline
\textbf{Example} & \multicolumn{2}{l}{\textit{\textbf{\begin{tabular}[c]{@{}c@{}}``Please help me to arrange a young woman \\ housekeeper with low price"\end{tabular}}}} \\ \hline
\textbf{Process} &  \textbf{Explanation} & \begin{tabular}[c]{@{}c@{}}\textbf{The result of} \\ \textbf{Each process}\\ \textbf{of the example}\end{tabular}   \\ \hline
\begin{tabular}[c]{@{}c@{}}Situational\\ judgment\end{tabular} & \begin{tabular}[c]{@{}c@{}}Determine \\whether the user \\chats or Q\&A\end{tabular} & \begin{tabular}[c]{@{}c@{}}A \textbf{\textit{real need}},\\ not a \textit{\textbf{chat}} \end{tabular} \\ \hline
\begin{tabular}[c]{@{}c@{}}Domain\\ identify\end{tabular} & \begin{tabular}[c]{@{}c@{}}Judge possible \\domain,\\ sort and record\end{tabular} & \{\textbf{\textit{Housekeeping}}, \textbf{\textit{Job}},...\} \\ \hline
\begin{tabular}[c]{@{}c@{}}Named\\ Entity\\ Recognition\end{tabular} & \begin{tabular}[c]{@{}c@{}}Extract entities from\\ user requirements\end{tabular} & \begin{tabular}[c]{@{}c@{}} $\{\{$pro:$\{ '\textbf{Housekeeper}'\}$,\\ price:$\{ '\textbf{low}'\}$, \\
	gender:$\{ '\textbf{woman}' \}$,\\ age:$\{ '\textbf{young}' \} \} \}$\end{tabular} \\ \hline
\begin{tabular}[c]{@{}c@{}}Search\\ in KG\end{tabular} & \begin{tabular}[c]{@{}c@{}}Find the point \\ and correlation of\\  user's requirement\end{tabular} & \begin{tabular}[c]{@{}c@{}}Corresponding\\ service  personnel\\ with their attributes\end{tabular} \\ \hline
\begin{tabular}[c]{@{}c@{}}Generate \\ intermediate\\ answer\end{tabular} & \begin{tabular}[c]{@{}c@{}}The answer obtained\\ by the query graph\\ is processed by \\the prepared answer \\ template and\\  returned to the user.\end{tabular} & \begin{tabular}[c]{@{}c@{}}``What are the\\ \textit{experience} restricts?"\end{tabular} \\ \hline
\end{tabular}
\vspace{1ex}
\end{table}
\subsection{NLU Module}\label{subsec:lp_trans}
\vspace{-0.5ex}
NLU Module performs quite different in the initial round and follow-up rounds of one conversation.
\subsubsection{Initial Round}\label{subsubsec:initial} This round happens after the bot is activated, and it would receive the first user command in terms of one short sentence. This command is always a fuzzy text requirement. As mentioned in Definition \ref{def:intrec}, this command input contains user initial intentions $D$, which can be recognized by the fine-tuned \textbf{BERT} model\footnote{This fine-tuned \textbf{BERT} model is trained based on the google BERT model (\url{https://storage.googleapis.com/bert_models/2018_11_03/chinese_L-12_H-768_A-12.zip}) with domain-specific corpus.}. 
If $D$ is empty, this command would be regarded as a chatting command, which cannot trigger the follow-up rounds. Otherwise, the intention set $D$ would be transferred into the reasoning module. Meanwhile, the result $\beta$ of the module ``domain identify” would be saved for follow-up rounds.


\subsubsection{Follow-up Rounds}

The reasoning module with initial intention set $D$ can generate the first reply (the detail would be explained in the next section). According to this reply, the user reacts with a new command. It would enter the module ``domain identify” function as well. If its output does not equal the saved $\beta$, then the bot considers that it a new dialogue. Otherwise, it is the follow-up round. It would repeat the tasks in the first round until the conversation ends. The user intention set $D$ would be filled up in every round. An example structure of $D$ and $H$ shows as follows:
\vspace{-1ex}
\begin{equation*}
\begin{split}
&H_{11}=pro:\{ 'Housekeeper'\}; H_{12}=price:\{'low'\}; \\
&H_{13}=gender:\{ 'woman' \}; \quad H_{14}=age:\{ 'young' \};
\end{split}
\end{equation*}
\vspace{-3ex}
\begin{equation*}
\begin{split}
D_{1} = \{ H_{11}, H_{12}, H_{13}, H_{14} \}; 
\end{split}
\end{equation*}
\vspace{-3ex}
\begin{equation*}
\begin{split}
D_{result} & = \{ D_{1} \} = \left \{ \left\{\; pro: \left\{ 'housekeeper' \right \} , price:\left \{ 'low' \right \} ,\right. \right. \\
& gender:\left\{ 'woman' \right\}, age:\left\{ 'young' \right\} \} \}
\end{split}
\end{equation*}

\vspace{-3.0ex}
\subsection{Reasoning Module}\label{subsec:reason}
\vspace{-1ex}
The Q$\&$A method designed in this paper is quite similar to the process of information retrieval and knowledge reasoning. It is based on a service knowledge graph as defined in Definition \ref{KG}. KG is responsible for searching or inferring the qualified candidate services according to the user requirements, so as to support the reply generation of each follow-up round. For example, \textit{``how about eating fried chicken at noon today?”.} Obviously, \textit{``fried chicken”} is the goal. KG has to help clarify this goal by identifying the following factors, timely store, satisfied price, proper delivery time, \textit{etc}. These factors are the attribute nodes of the KG. Another example, \textit{``I prefer something warm to make my stomach comfortable."} In this case, the goal is missing. We have to firstly infer the goal with those presented attributes based on the KG.
As shown the Reasoning Module in Fig. \ref{module2}, information retrieval was processed through the connections between entities in KG.
\subsubsection{Mapping Concept to Knowledge Graph Entities} \quad
\par{As shown in Fig. \ref{reason}, we match the concepts identified in $D_{result}$ to entities for reasoning in KG. The system generates lists of proper nouns when creating the KG. When restricts whose label is $pro$ are identified in $D$ (Profession is the main entity intention in the demand of human services by default), reasoning module maps the ``HouseKeeper" to the entity in KG by looking up the list and matching ($purple, ID=336$).}
\subsubsection{Reasoning in Konwledge Graph} \quad
\par
\begin{figure}[htbp]
\centering
\includegraphics[width=0.9\linewidth]{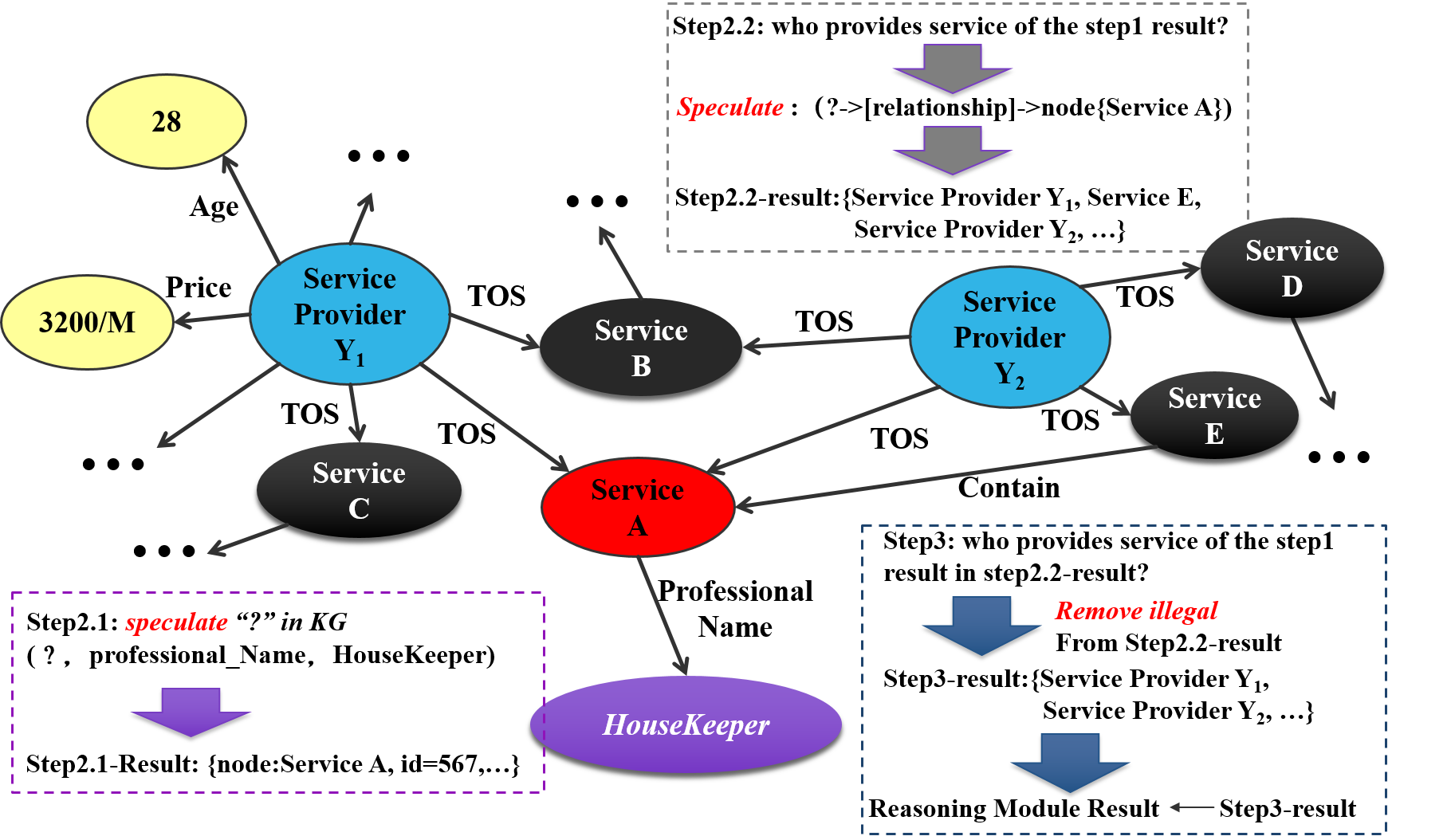}
\caption{Reasoning Module}
\label{reason}
\vspace{-1ex}
\end{figure}

{\textbf{Mapping entity concepts to services} As step2.1 shows in Fig. \ref{reason}, the corresponding service entity ($red, ID=567$) is found through the representational learning method, such as Trans \cite{wang2014knowledge}, which can help to infer the entire connected entities (relevant entities).} 
\par{\textbf{Tracing the services to service providers} The set $T$ obtained by step2.2 is the set of entities connected with the entity (\textit{red}) in the result of step2.1. (\textit{Two Y entities in blue and One service entity E in black} who may capable of providing this service A.)}
\subsubsection{Result Filtering}
Since the user requirement must be finally satisfied by one specific service provider,
$sn_{i}$ in the result $T$ of step2.2 must be either the entity with label \textit{per} or the entity connected with the one labeled \textit{per} ($sn_{i}-[employ] \rightarrow per$).
Thus, the $sn_{i}$ that is not qualified with these constraints would be removed from $T$. The refined $T$ of step3 is the output of the reasoning module.
\vspace{-1.5ex}
\subsection{Dialogue Management Module}\label{DMM}
\vspace{-0.5ex}
\label{subsec:diaman}
This module has two parts, pruning strategy and dialog policy. Pruning strategy take in charge of an offline task to generate the conversation rules based on the system owned data. These rules would support the dialog policy module to lead the conversation round by round. Because the services have tremendous attributes, if the bot confirms the requirement by inquiring every attribute, then the multi-rounds conversation could be a mess. In order to avoid this, this paper proposes a granular computing (GrC) based method to cluster these attributes as defined in Definition \ref{PRUNE}. This pruning method would provide a proper inquire path as the dialog policy. 
\subsubsection{Offline Before Dialogue Beginning}
GrC method takes the various types of service combinations as the root data $C$ for the pruning as Fig. \ref{decision_tree}. GrC divides data into optimal granules $C=\left\{\boldsymbol{C}_{1}, \boldsymbol{C}_{2}, \boldsymbol{C}_{3}, \cdots, \boldsymbol{C}_{p}\right\}$ once the value of $Goal$ defined in Definition \ref{PRUNE} tends to be stable. The current round inquiring attribute set was calculated by analyzing the clustering center $m_{i}$. The system traverses every cluster class, depending on all available choices by users (different granules in GrC result), then processes the next round GrC with the result set after pruning in each round, until the number of leaf node elements clustered by GrC is less than a threshold value $N$.
The $N$ is a parameter to be tuned, which determines the number of the most suitable leaves return to the user.
$N$ can't be too small, because not all kinds of data sets can be of excellent particle size. But at the same time, $N$ cannot be too large either, otherwise, it would lose the essence of service recommendation. The system generates $N$ by algorithm \ref{calN}.
$X$ is the upper limit of Service Recommendation at a time, and $Res$ is the GrC algorithm result array, which includes the number of services contained in each leaf node in the result of GrC pruning. 

\begin{algorithm}\small
\caption{Calculate N in GrC algorithm automatically}  
\begin{algorithmic}[1] 
\Require Upper limit $X$; GrC algorithm result array $Res$
\Ensure $N$ 
\State $Res\_fre \gets \{\}$
\For {each different value in $Res$}  
\State $Res\_fre \gets Res\_fre +(value,frequency)$
\EndFor
\State $frequency\_in\_order \gets$ sort each pair in $Res\_fre$ in $value$'s ascending order
\State $M \gets$ The serial number of the median of the element in $frequency\_in\_order$, $Candidate \gets \{\}$
\For {$p = M \to len(Res\_fre)-1$}  
\State $Canddiate \gets Candidate + frequency\_in\_order[p]$
\EndFor
\State $max\_pair \gets$ the pair where have the max frequency in $Candidate\footnote{fetch the larger value's pair when there are the same frequency}$, $v \gets 0$, $i \gets$ The serial number of $max\_pair$ in $Candidate$
\While{$i<len(Candidate)$ \textbf{and} $v < X$}  
\State $v \gets Candidate[i].value$   
\If{$i = 0$}
\State $\Delta_{left} \gets infinite$
\State $\Delta_{right} \gets Candidate[i].frequency-$ 
\Statex $\qquad \qquad \qquad Candidate[i+1].frequency$
\ElsIf{$i = len(Candidate)-1$}
\State $\Delta_{right} \gets infinite$
\State $\Delta_{left} \gets Candidate[i].frequency-$   
\Statex $\qquad \qquad \qquad Candidate[i-1].frequency $
\Else
\State $\Delta_{right} \gets Candidate[i].frequency -$ 
\Statex $\qquad \qquad \qquad Candidate[i+1].frequency $
\State $\Delta_{left} \gets Candidate[i].frequency -$ 
\Statex $\qquad \qquad \qquad Candidate[i-1].frequency $	
\EndIf
\State $\delta_{left} \gets \vert \Delta_{left} \vert$, $\delta_{right} \gets \vert \Delta_{right} \vert$
\If{$\Delta_{right}<0$ \textbf{and} $\Delta_{left}<0$}
\If{$\delta_{left}$ \textless $\delta_{right}$}
\State $v \gets  Candidate[i-1].value$
\State \textbf{break}
\EndIf
\ElsIf{$\Delta_{right}>0$ \textbf{and} $\Delta_{left}>0$}
\If{$\delta_{left}$ \textgreater $\delta_{right}$}
\State \textbf{break}
\EndIf
\ElsIf{$\Delta_{right}>0$ \textbf{and} $\Delta_{left}<0$}
\If{$\delta_{left}$ \textgreater $\delta_{right}$}
\State $v \gets Candidate[i-1].value$
\EndIf
\State \textbf{break}
\EndIf
\State $i \gets i+1$
\EndWhile
\State $N \gets v$  
\end{algorithmic}
\label{calN}  
\end{algorithm}

And for the selection of class clusters number $p$, the system quotes \textit{fpc} index. \textit{fpc} is fuzzy partition coefficient, which is an index to evaluate the classification. It ranges from 0 to 1, and 1 works best. The algorithm sets $p$ at $2 \leq p \leq \sqrt{n}$, tests \textit{fpc} values under different $p$, and selects $p$ corresponding to the maximum \textit{fpc} value as the default number of $p$.

The dialog policy that means the inquiring attribute in every round is determined based on \textbf{\textit{data}}. Dialog policy offline in DMM can avoid heavy computation online effectively and still speed up the dialogue process (Reduce the number of dialogue rounds).
\subsubsection{Conversation Process Online}
After dialog policy confirming the inquiring attribute of each round, user chooses proper granules $C_{i}$ belongs to the result of the GrC pruning $C=\left\{\boldsymbol{C}_{1}, \boldsymbol{C}_{2}, \boldsymbol{C}_{3}, \cdots, \boldsymbol{C}_{p}\right\}$. DMM accepts user feedback and determines the corresponding solution path in the GrC result tree until the candidate set $T$ achieve a leaf node. DMM formulates a data-based dialog policy for the fastest and most efficient elicitation of user requirement services by the GrC method. The dialog policy in DMM can help the system elicit user requirements and accelerate conversation end in the least round.
\subsection{NLG Module}\label{subsec:nlg}
\vspace{-0.5ex}
The purpose of the NLG module is to improve the interactivity between users and the system. 
The module accepts input in a non-verbal format and converts it into human-readable format sentences. 
When conversational state receives the output data by the dialogue management module, it determines whether the output conforms to the termination state $EndTag$. 
If output meets the end conditions, the NLG module receives the DMM output data and transfers the human-readable sentences as the final result to the user; otherwise, 
the system would generate the intermediate results and return to the user, waiting for user's feedback as the TABLE \ref{NLG} below. 

\begin{table}[htbp]
\vspace{-3.5ex}
\scriptsize
\centering
\caption{Example of NLG Module Result (N = 8)}
\label{NLG}
\begin{tabular}{cccl}
\hline
\begin{tabular}[c]{@{}c@{}}\textit{\textbf{Back-End}} \end{tabular} & \begin{tabular}[c]{@{}c@{}}\textit{\textbf{End}}\\ \textit{\textbf{Tag}}\end{tabular} & \begin{tabular}[c]{@{}c@{}}\textit{\textbf{Element}} \\ \textit{\textbf{Quantity}} \\ \textit{\textbf{in Set}} \end{tabular} & \multicolumn{1}{c}{\textit{\textbf{Return to User}}} \\ \hline
\begin{tabular}[c]{@{}c@{}}Tag=\{`Price'\}\end{tabular} & 0 & 1 & \multicolumn{1}{c}{What are the experience restricts?} \\ \hline
\begin{tabular}[c]{@{}c@{}}Id=\{'386',\\ '624', '125',\\$\cdots$  '444'\}\end{tabular} & 1 & 9 (\textgreater8) & \begin{tabular}[c]{@{}l@{}}No attributes left and we get \\a lot of services for you:\\ \{1:\{Name:'Lily',Age:'23', Price:'3200',…\};\\
2:\{Name:'Rose',Age:'22',Price:'3500',…\};\\     \ \  3:\{Name:'Lisa',Age:'25',Price:'2400',…\};\\ …\}\end{tabular} \\ \hline
\begin{tabular}[c]{@{}c@{}}Id=\{'586',\\ '633','636'\}\end{tabular} & 1 & 3 ($\leq$8) & \begin{tabular}[c]{@{}l@{}}Prepare three services for you:\\ \{1:\{Name:'Amy',Age:'28',Price:'2700',…\};\\    2:\{Name:'Fred',Age:'28',Price:'2400',…\};\\     3:\{Name:'Anda',Age:'27',Price:'2600',…\};\\ \}\end{tabular} \\ \hline
\end{tabular}
\vspace{-1.5ex}
\end{table}

\subsubsection{Intermediate Q\&A}
When conversational state judges dialogue management module output does not satisfy the conditions that no attribute deserved to be classified  or the number of the results is small enough, the system generates intermediate inquiry base on the template where templates and grammar are rule-based strategies to finish multi-round dialogue NLG module. The system displays the output through modules defined in advance. Take elderly services as an example, and the sentence is dynamically changed and generated by a predefined set of business rules 
(such as the if/else loop statement). 
The return to user of first line in table is an inquiring sentence based on module input \textit{\{Tag=\{Price\};End\_Tag=0;Quantity=1\}}.

\subsubsection{Final Answer Generation} 
When the end conditions are met, the system would execute the Answer Generation function. Whatever $N$ is in the final $T$,
the system would return the complete information of the first $N$ possible solutions to the user for selection in descending order of user matching finally. 
As the example in TABLE \ref{NLG}, when \textbf{End\_Tag} meets $1$, NLG generates readable return which is the detailed information of service providers in \textbf{Back-End} input as the final result to the user. If the quantity of the element in the set is more than $N$, the return sentence would show the first $N$ service providers and point out there are no attributes left. Otherwise, NLG only displays the corresponding information as the final result. 
\vspace{-1ex}
\section{Experiments and Results}
\vspace{-0.5ex}
\subsection{Experiment Setup}\label{experimentsetup}
\vspace{-0.5ex}
\textbf{DataSet} 
One of the data sets is used to fine-tune and validate the pretrained Google BERT model. This data set\footnote{\url{https://github.com/tian231825/Conversational_AI_Bot/tree/master/project/BERT/model_data}} has been divided into train\_set, dev\_set, and test\_set in 8:1:1. Finally, the accuracy of this model has achieved 89.5\% based on 1975 valid corpus, better than BiLSTM\_Attention 84.7\%.
Another data set\footnote{\url{https://github.com/tian231825/Conversational_AI_Bot/tree/master/data}} is used to construct the service knowledge graph. This data set contains 827 service providers. Every service has more than 9 attributes. After removing the illegal or missing data, the constructed KG has 9478 triples, which includes 960 entities and 10 relationships.
\par
\begin{figure}[htbp]
\vspace{-2.0ex}
\centering
\includegraphics[height=4.7cm,width=8cm]{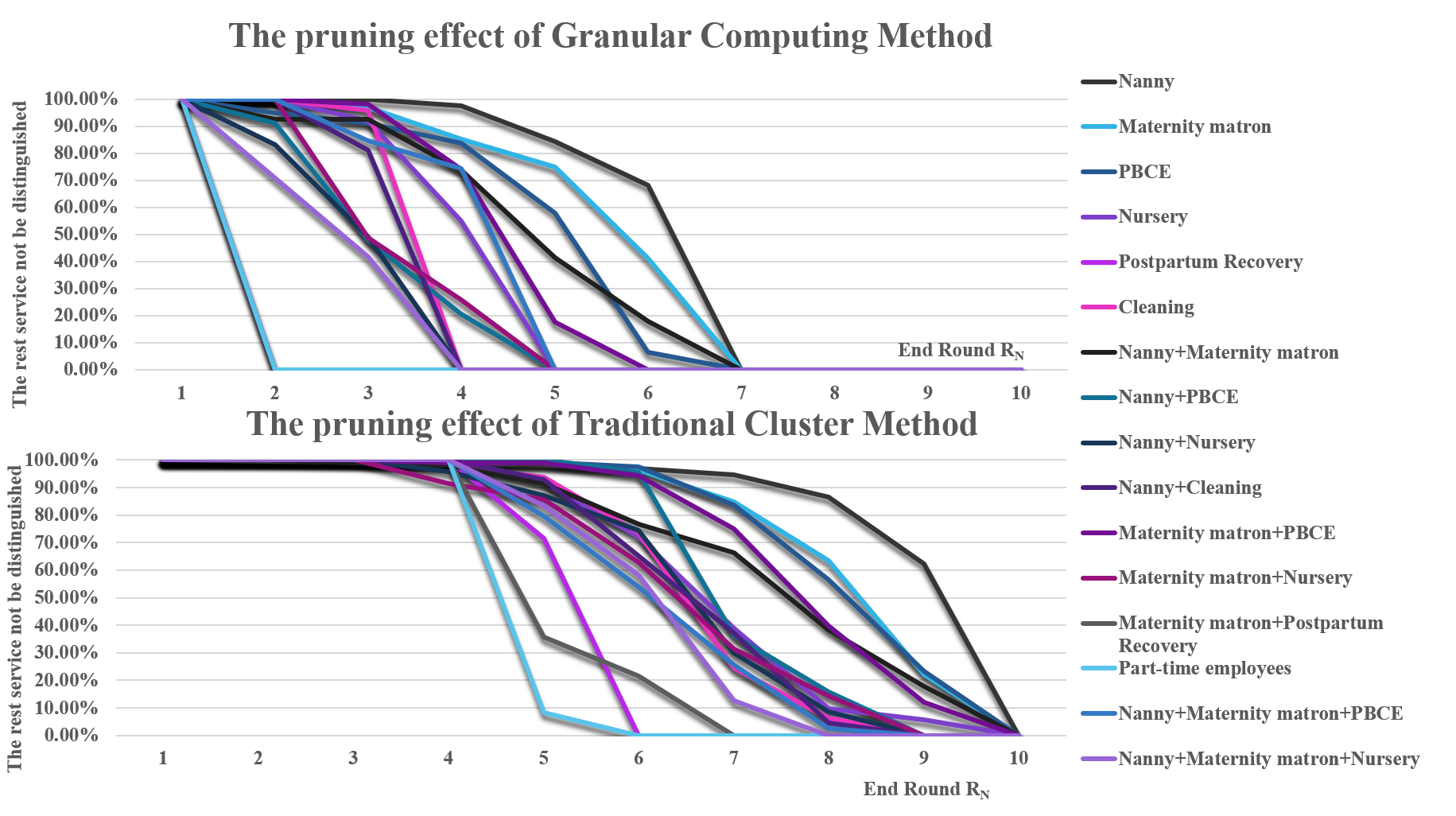}
\vspace{-1.5ex}
\caption{The pruning effect of Two Cluster Method}
\label{trad_clu}
\vspace{-1ex}
\end{figure}

{\textbf{Baseline and Evaluation}
The experiment defines the k-means method as a traditional pruning strategy to simulate the whole process of multi-round dialogue. And the system uses FCM algorithm as an implementation of Grc method. Within the scope of the knowledge graph information, user requirement can be simulated as the generated user input to be accepted by the system. The NLU module loaded BERT, and the reasoning module would traverse and simulate all possible valid multi-round dialogue process, record the path generated by each decision.
The experiment judges the accuracy of the two methods by the $hit\_rate$ index. Users need and only need the best $1$ service by default. And $hit\_rate$ refers to the probability that our best service target would appear in the case of $N$ recommended items at one time calculating through conditional probability, such as equation \ref{goal}, and $l$ is the number of elements in leaf node:
\begin{gather}\label{goal}
Hit\_rate=\frac{C_{1}^{1} C_{l-1}^{n-1}}{C_{l}^{n}}=\frac{C_{l-1}^{n-1}}{C_{l}^{n}}
\end{gather}
Then the experiment statistics the end round number $R_{N}$ and calculates the average round through $\frac{\sum_{1}^rR_{N}}{r}$, $r$ is the number of test cases.
The experiment result compares two methods by $hit\_rate$ and average round. The two cluster results of 16 types of service combinations are shown in Fig. \ref{trad_clu}.}
\vspace{-1ex}

\subsection{Result and Analysis}\label{subsec:resandana}
In TABLE \ref{tab:hit}, the experimental results indicate that the recommendation accuracy decreases with the decrease of rounds. In practical application, the paper is more inclined to make the dialogue rounds and final candidate set accurahcy both achieve better results, not just one to achieve the optimal. We believe that a 36.1\% decreasing in the average rounds would make the conversational AI bot user experience better than a 1\% decreasing in accuracy.

\begin{table}[ht]
\scriptsize
\vspace{-3ex}
\centering
\caption{Average Hit Rate of two method}
\label{tab:hit}
\begin{tabular}{ccc}
\hline
\textit{\textbf{}} & \textit{\textbf{\begin{tabular}[c]{@{}c@{}}Traditional Cluster\\ Method\end{tabular}}} & \textit{\textbf{\begin{tabular}[c]{@{}c@{}}Granular Computing \\ Method\end{tabular}}} \\ \hline
HIT rate(\%) & 95.06 & 94.79 \\ \hline
Avg\_round & 8.391 & 5.357 \\ \hline
\end{tabular}
\vspace{-2ex}
\end{table}

Fig. \ref{trad_clu} also shows the pruning strategy efficiency of all service types based on granular computing. The experiment result can get a conclusion from the figure: the GrC method is universal and not only valid for certain types of data. In contrast to the figure of the traditional cluster method, you can see that the GrC method is indeed effective in decreasing the round of each type of service screening process.
\begin{figure}[htbp]
	\vspace{-2ex}
	\centering
	\includegraphics[height=3cm,width=4.0cm]{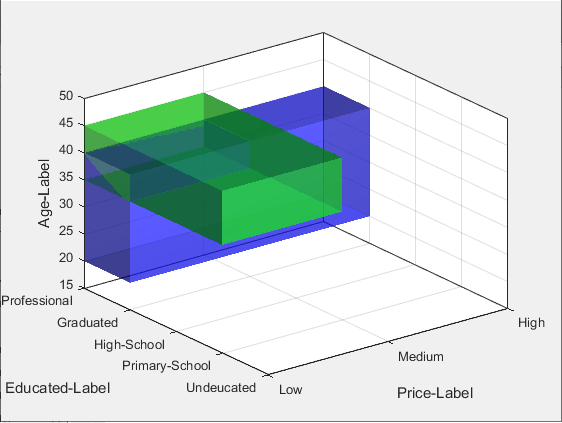}
	\caption{Example of GrC solution space}
	\label{GrC}
	\vspace{-2ex}
\end{figure}

At the same time, the paper is pleased to find that GrC method solves the problem of the fuzzy boundary of continuous variables.
For example, continuous \textit{``price"} can be classified according to data automatically rather than manual operation. As shown in Fig. \ref{GrC}, the system can accept the fuzzy requirement like \emph{``low price"} in TABLE \ref{process}. Traditional methods are disadvantageous for dealing with such fuzzy requirements. If the price is between 0 and 4000, and what does \textit{``expensive"} mean? Artificial can define more than a particular value means \textit{``expensive"},such as \textit{``3000"}. But it's hard to say \textit{``2999"} or \textit{``2998"} is medium or cheap although they are lower than \textit{``3000"}. GrC method can solve this, and after data divided into several general categories, the system only needs to get the entities in the corresponding solution space according to user requirement. Fig. \ref{GrC} shows that two requirements for the staff.
\vspace{-3ex}

\begin{figure}[ht]
\centering
\subfigure[Classified by Age]{
\begin{minipage}[t]{0.29\linewidth}
	\centering
	\includegraphics[width=\textwidth]{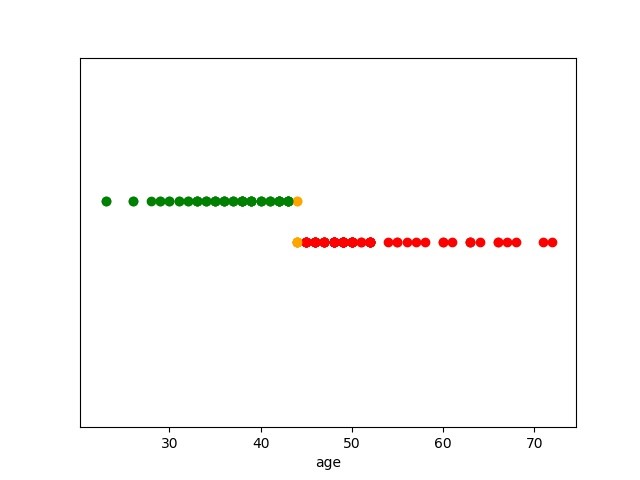}
	\label{fig:1}
\end{minipage}
}
\vspace{-1ex}
\centering
\subfigure[Classified by Working Experience]{
\begin{minipage}[t]{0.29\linewidth}
	\centering
	\includegraphics[width=\textwidth]{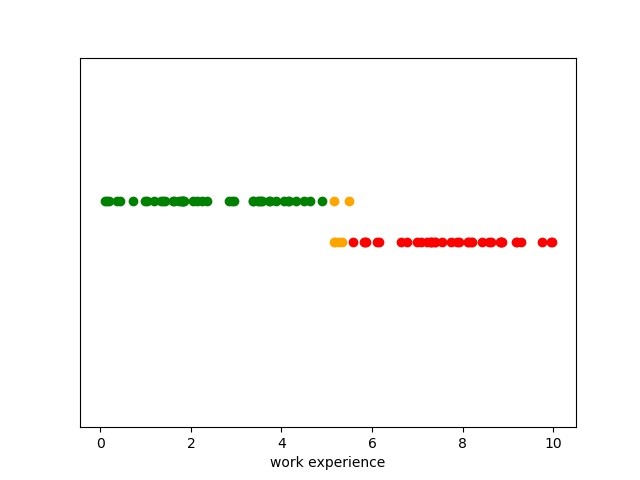}
	\label{fig:2}
\end{minipage}
}
\centering
\subfigure[Classified by Educational Background and Service Area]{
\begin{minipage}[t]{0.29\linewidth}
	\centering
	\includegraphics[width=\textwidth]{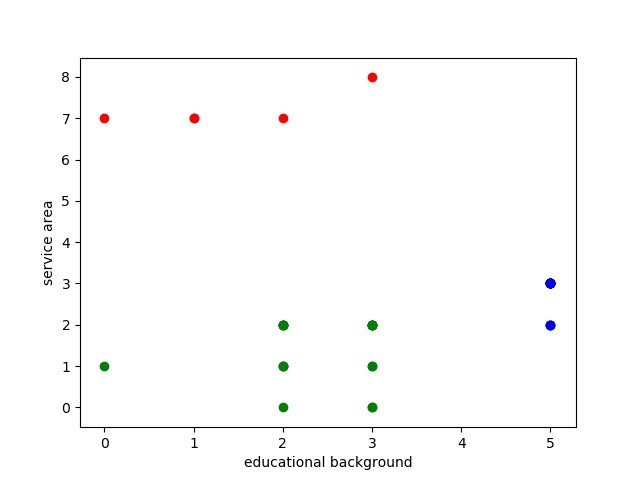}
	\label{fig:3}
\end{minipage}
}
\centering
\subfigure[Classified by User Evaluation]{
\begin{minipage}[t]{0.29\linewidth}
	\centering
	\includegraphics[width=\textwidth]{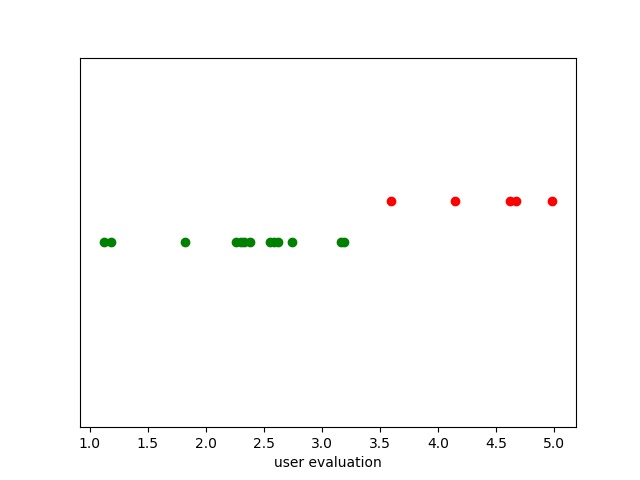}
	\label{fig:4}
\end{minipage}
}
\centering
\subfigure[Classified by Service Area]{
\begin{minipage}[t]{0.29\linewidth}
	\centering
	\includegraphics[width=\textwidth]{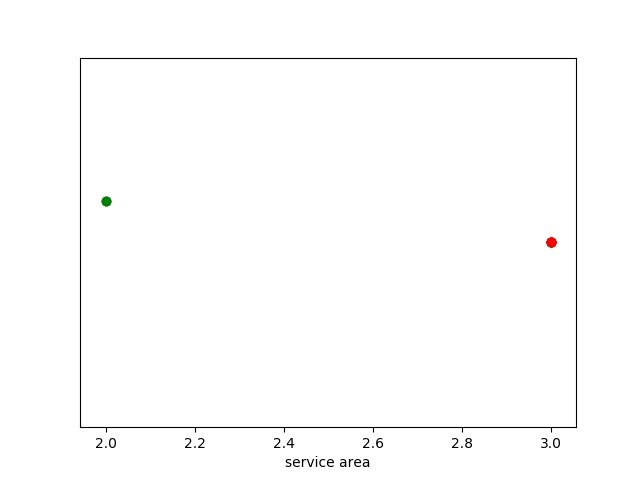}
	\label{fig:5}
\end{minipage}
}
\centering
\subfigure[Classified by Service Price]{
\begin{minipage}[t]{0.29\linewidth}
	\centering
	\includegraphics[width=\textwidth]{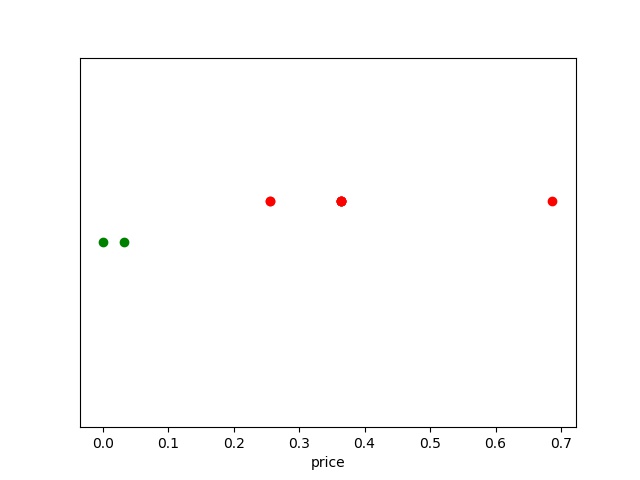}
	\label{fig:6}
\end{minipage}
}
\caption{Service PBCE pruning process in the system.}
\label{process_fig}
\vspace{-1ex}
\end{figure}

\begin{figure}[htbp]
\vspace{-2ex}
\centering
\includegraphics[width=\linewidth]{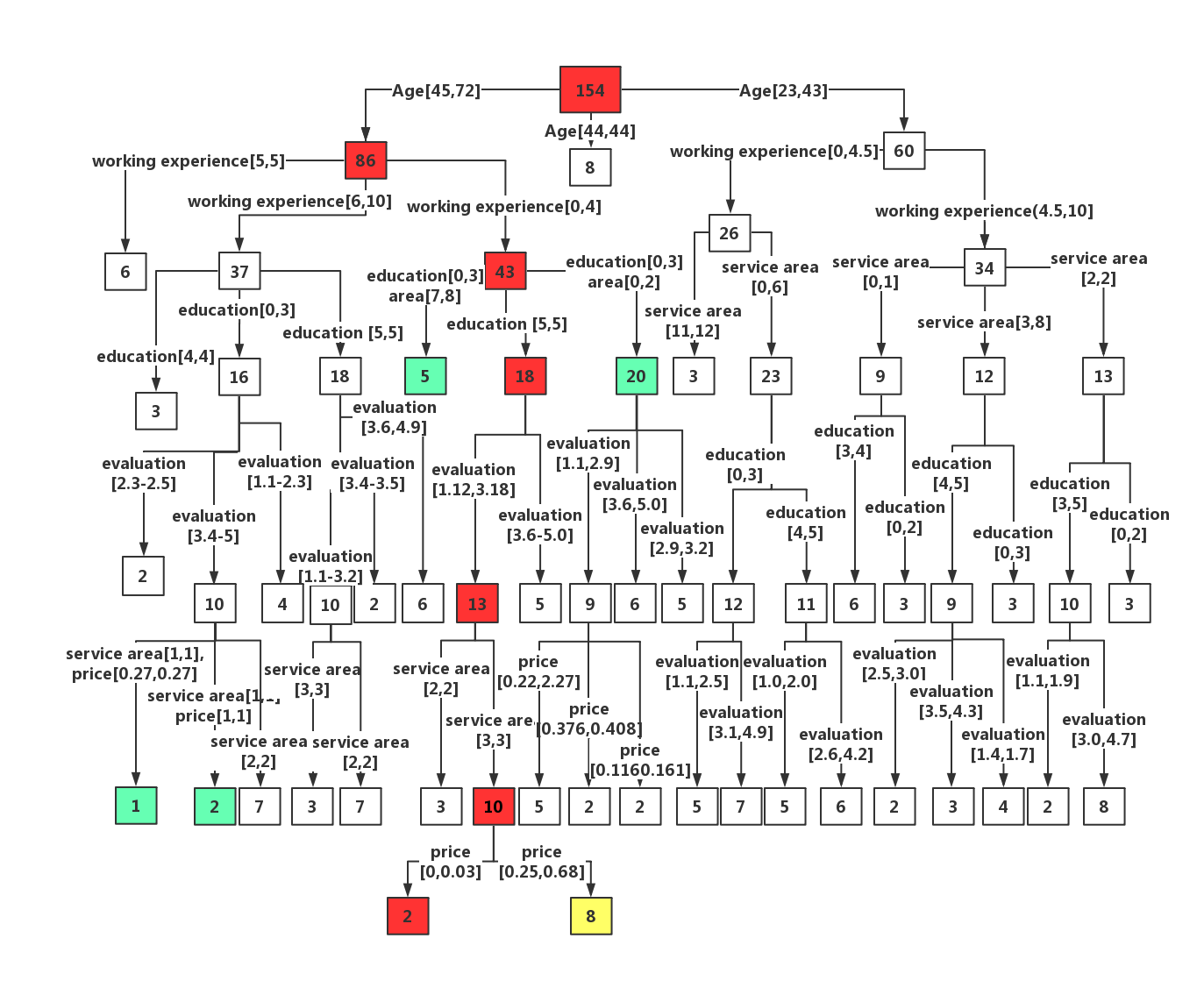}
\vspace{-5ex}
\caption{The Result Tree of Service PBCE from GrC Method}
\label{decision_tree}
\vspace{-1.0ex}
\end{figure}

The six graphs in Fig. \ref{process_fig} are the granular calculation steps in the dialog process that simulates the service type PBCE (nursery teacher). The experiment normalized the service price, service provider educational background, and service area to display better in the results. In Fig. \ref{fig:1},(b),(d),(e),(f), we can notice that the Y-axis in the figure has no value because the figures uses a two-dimensional projection to make the overlapping part of the granules visible. And the service set is constantly dismembered until the end of Fig. \ref{fig:6} in the process. It took up at most seven rounds dialogue (the elicitation of the service type is seen as one round dialogue). The red path in Fig. \ref{decision_tree} has shown the process of Fig. \ref{process_fig}. Furthermore, Fig. \ref{fig:1} shows the second round result, it can be observed that the age-based division is such that the data set can aggregate into two granularity with distinct boundaries. And the same situation in Fig. \ref{fig:2}, the system gathered the overlapping parts of the classification results as a single granularity. In the two results in Fig. \ref{fig:3} and Fig. \ref{fig:5}, since the separate attribute belongs to discrete non-continuous variables, the figures show that the data set coincide at some points, and the algorithm automatically divides the granules according to the result. In the fourth round results Fig. \ref{fig:3}, the granular calculations divide the data into several categories in 1 round dialogue through multidimensional attributes, which is why the total round number of the GrC method is much lower than the traditional cluster method. Divided by more than one attribute also occurs at the green point in Fig. \ref{decision_tree}. And the paper marks the longest dialog path with red points.

\begin{figure}[htbp]
\vspace{-1ex}
\centering
\includegraphics[width=0.9\linewidth]{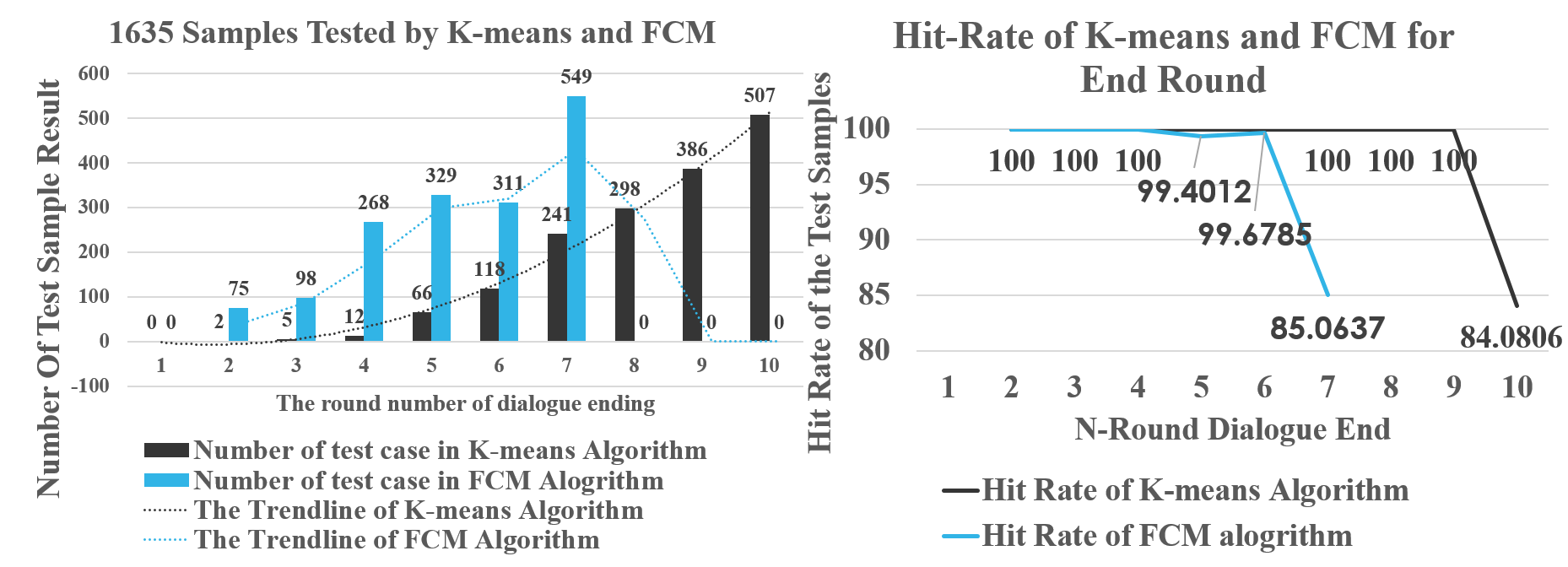}
\vspace{-1.5ex}
\caption{Comparision of Two Methods}
\label{all_compare}
\vspace{-1ex}
\end{figure}

The Fig. \ref{all_compare} shows that the k-means algorithm early-round hit ratio is 100. This is because the strategy system adopted is to stop clustering until the leaf node is less than 8 ($N=8$). If the leaf node is more than 8, the system would continue to the next round. Therefore, there would be no leaves greater than 8 (the available attributes are no longer available) until the 10th round, resulting in a hit ratio below 100. The actual comparison should be the hit ratio of the two methods when the available attributes are no longer available. That is, the hit ratio of the FCM method in the seventh round (85.0638) and that of the k-means method in the tenth round (84.0806).

This paper compares the results of the two methods in Fig. \ref{all_compare}. The highest duration simulation dialogue of the GrC method ends in the seventh round.  Compared to the traditional cluster method, GrC method can directly aggregate some of the services into a single granule multidimensional instead of another questioning from the second round. 

We can get definite conclusions from the comparison. The GrC-based algorithm is superior to the traditional clustering algorithm in the formulation of decision strategies. 
\section{Conclusions and Future Work}
\vspace{-0.5ex}
This paper proposes a user intention recognition method for conversational AI bot based on the finetuned BERT model with human services related data. The requirement elicitation procedure is a multi-round conversation leading by conversational AI bot dynamically, rather than predefined rules. During this, the bot can recognize the user's intention through not just one-by-one analysis of the single round. Thus, the bot can identify the fine-grained requirements in a continuous context. A decision-making pruning method based on granular computing is proposed to avoid template matching user input. It can effectively deal with the fuzzy requirement concept. The experimental results show that compared with the traditional clustering method, our method can effectively reduce the conversation rounds and ensure the user experience. Meanwhile, the excessive hit rate of the algorithm remains quite high.

Conversational AI bot is still very challenging to achieve leapfrog development in the future. Massive domains related to data, even the crossover domains data, are further required to remain or enhance the intelligence and effectiveness of this conversational service delivery method. Honestly, base on the method proposed in this paper, underlying more in-depth requirements would not be recognized very clearly such as the requirements need to elicited by \textit{multi-hop in the text}, and we still need to make the pruning strategy offline before the process, and we hope to be able to make it online in real-time in the future. These issues mentioned above would be our future works.

\section*{Acknowledgement}
Research in this paper is partially supported by the National Key Research andDevelopment Program of China (No 2018YFB1402500), the National ScienceFoundation of China (61802089, 61832004, 61772155,  61832014).
\vspace{-1ex}
\newcommand{\BIBdecl}{\setlength{\itemsep}{0.001 em}}

\bibliography{references}



\end{document}